\documentclass[5p,times]{elsarticle}

\usepackage{amsmath}            
\usepackage{epstopdf}           
\usepackage{flushend}           

\usepackage{subcaption}
\usepackage{amssymb} 
\usepackage{lineno,hyperref}
\usepackage{algpseudocode}
\usepackage{amssymb}
\usepackage{amsmath}
\usepackage{siunitx}
\usepackage{tabularx}
\usepackage{multirow}
\usepackage{amsthm}
\usepackage{graphicx}
\usepackage{lscape}
\usepackage{xspace} 

\begin{document}

\begin{frontmatter}
 
\title{Large language model for Bible sentiment analysis: Sermon on the Mount}

\author[Second]{Mahek Vora} 
 
\author[Third]{Tom Blau}
 
\author[Fourth]{Vansh Kachhwal}

\author[Fifth]{Ashu M. G. Solo}

\author[First,ping]{Rohitash Chandra \corref{cor1} }
\ead{rohitash.chandra@unsw.edu.au}

\address[Second]{  Indian Institute of Technology Guwahati, Assam, India}

\address[Third]{Data61, Commonwealth Scientific and Industrial Research Organisation, Sydney, Australia}

\address[Fourth]{Indian Institute of Technology Delhi, India}

\address[Fifth]{ Maverick Trailblazers Inc., Wilmington, Delaware, U.S.A.}
\address[First]{ Transitional Artificial Intelligence Research Group, School of Mathematics and Statistics, UNSW Sydney, Sydney, Australia}

\address[ping]{Pingala Institute of Artificial Intelligence, Sydney, Australia}

\begin{abstract}

The revolution of natural language processing via large language models has motivated its use in multidisciplinary areas that include social sciences and humanities and more specifically, comparative religion. Sentiment analysis provides a mechanism to study the emotions expressed in text. Recently, sentiment analysis has been used to study and compare translations of the Bhagavad Gita, which is a fundamental and sacred Hindu text.  In this study, we use sentiment analysis for studying selected chapters of the Bible.  These chapters are known as the Sermon on the Mount. We utilize a pre-trained language model for sentiment analysis by reviewing five translations of the  Sermon on the Mount, which include the King James version, the New International Version, the New Revised Standard Version, the Lamsa Version, and the Basic English Version. We provide a  chapter-by-chapter and verse-by-verse comparison using sentiment and semantic analysis and review the major sentiments expressed. Our results highlight the varying sentiments across the chapters and verses. We found that the vocabulary of the respective translations is significantly different.  We detected different levels of humour, optimism, and empathy in the respective chapters that were used by Jesus to deliver his message.

\end{abstract}

\begin{keyword} 
  BERT, natural language processing, comparative religion, Bible, sentiment analysis 


\end{keyword}

\end{frontmatter}

\section{Introduction}



 Natural language processing (NLP) enables computers to read, interpret, translate, analyze, and understand written and oral forms of language \cite{indurkhya2010handbook,manning1999foundations,chowdhury2003natural}.   NLP includes tasks such as language translation, speech recognition, topic modelling, and semantic and sentiment analysis \cite{manning1999foundations} and is typically implemented using machine learning and  deep learning \cite{
otter2020survey,cambria2014jumping,alshemali2020improving}.  Sentiment analysis of written and oral text provides an understanding of human emotions and affective states  \cite{liu2012survey,medhat2014sentiment,hussein2018survey}. Semantic analysis has a wide range of applications that mainly  include marketing with applications in recommender systems in social media \cite{ray2021ensemble,rosa2018knowledge}.  Semantic analysis has been used in market research to determine customer viewpoints about products or services  in\cite{aggarwal2009using, rambocas2017online}.  Sentiment analysis has a wide range of social media applications \cite{yue2019survey} that can guide sociologists, political scientists, and policymakers. Sentiment analysis has been used to study public behaviour and voting preferences during the 2020 United States presidential elections \cite{chandra2021biden}.  There  is also research on sentiment analysis from social media during COVID-19 \cite{nemes2021social} and vaccine-related COVID-19 sentiments 
In a recent study, Chandra and  Kulkarni \cite{chandra2022semantic} used  sentiment analysis  to study and compare translations of the sacred Hindu text known as the Bhagavad Gita. The study reported that although the style and vocabulary of the translations differ substantially, the semantic and sentiment analysis showed similarity in the majority of the verses.  The study also showed how sentiment analysis captures the changing topics of the conversation between the protagonists, Lord Krishna and Arjuna, over time. Chandra et al. \cite{shukla2023evaluation} used sentiment analysis to compare the Sanskrit translation of the Bhagavad Gita with expert (human) translators and reported major discrepancies in Google Translate. These applications have motivated us to use sentiment analysis in this study of selected chapters of the Bible.

The Bible is one of the most influential texts in history and has attracted an immense volume of scholarship. \cite{levenson1993hebrew,moorey1991century,young1999biblical} The field of biblical scholarship also has a  history of using non-computational linguistic tools and theory to study the structure, meaning, authorship, and authenticity of different parts, versions, and translations of the Bible~\cite{hamilton1903historical, haupt1925philologic,brown1927english,nida1972implications}. Artificial intelligence and computational linguistics have also been applied to the study of the Bible. This began with rule and grammar-based methods of computational linguistics~\cite{jappinen1986associative} and continues with modern deep learning-based  NLP  ~\cite{jones2022machine}. NLP serves as an ideal resource for the study of  the Bible and related texts  because they are large and highly structured and have been translated into hundreds  of languages  with most versions in the public domain.  The large and diverse set of English language translations raises the need for comparisons between different styles of translations to verify if they feature similarity in the core meaning \cite{carlson2018evaluating}. At the same time, the availability of the same text in a wide range of languages supports multilingual machine translation and provides a toehold for NLP in low-resource languages~\cite{kim2015part}. There are many language pairs where the Bible is one of the major texts available in both languages~\cite{agic2015if}. 

 The use of NLP has the benefit of saving a lot of time that it would take for humans to do a verse-by-verse comparison. Although the Bible has been used extensively as a resource for training and evaluating NLP models, applying NLP for understanding biblical scholarship is comparatively less common. In particular, although some prior works perform semantic analysis or topic modelling on the Bible, none of them employed novel deep learning models for NLP such as the sentiment and semantic analysis of the Bhagavad Gita~\cite{chandra2022semantic} and topic modelling between the Bhagavad Gita and the Upanishads~\cite{chandra2022artificial}.  These applications demonstrate the effectiveness of deep learning-based  NLP  in advancing our understanding of sacred texts. In spite of this, comparatively little work has been done to apply these tools to biblical scholarship. We aim to close this gap by using novel deep learning-based NLP models such as the \textit{bidirectional encoder representations from transformers} (BERT) \cite{devlin2018bert,rogers2020primer}, which has been prominently used for language modelling for a wide range of tasks that include sentiment analysis ~\cite{chandra2022semantic}.

In this study, we focus on selected chapters of the  Bible, which are known as the \textit{Sermon on the Mount.} Our goal is to use sentiment analysis to compare selected translations of the Sermon of the Mount including the King James version, the New International Version, the New Revised Standard Version,  the Lamsa Version, and the Basic English Version. We compare the vocabulary of the different translations using bigrams and trigrams to give us an understanding of the style used in the respective translations. 
 We provide a verse-by-verse comparison using sentiment and semantic analysis and review the major sentiments expressed. We visualise the polarity score of the verses and the chapters and evaluate the similarities and differences between the respective translations.



 The rest of the research paper is organised as follows: Section 2 provides background on related topics, Section 3 presents the methodology, and  Section 4 presents the results. Finally, Section 5 provides a discussion, and Section 6 concludes the paper.

 \section{Background} 
 
 \subsection{The Bible} 

 The Bible is a collection of religious texts by different authors that are considered sacred in Christianity, Judaism, and other religions. The texts in the Bible include stories, poetry, parables, instructions, and prophecies \cite{levenson1993hebrew,moorey1991century,young1999biblical}. The word "Bible" can refer to the Hebrew Bible \cite{} and the  Christian Bible \cite{}. The Christian Bible is subdivided into the Old Testament \cite{} and the New Testament \cite{}. The contents of the Old Testament vary between different Christian traditions, but it mostly overlaps with the Hebrew Bible and differs primarily in the order of books \cite{}. The Old Testament contains the Judeo-Christian cosmogony, the history of the Israelites (people from Israel), and Mosaic law~\cite{lightner1911mosaic} (commandments by Moses). The New Testament describes the life and teachings of Jesus and the Christian church. Millions of people around the world were converted to Christianity. A significant portion of indigenous populations  were forced to convert through invasions \cite {forcedconversions} and their descendants naturally became Christians. According to Guinness World Records, the Christian Bible is the best-selling book of all time with over 6 billion copies sold \cite{guinness}. Hence, the Bible has a profound influence on human history and culture. Thus, it is the focus of our study. 
 
 The Sermon on the Mount is widely regarded as the definitive summation of Jesus's ethical and spiritual teachings~\cite{davies1966sermon} and has consequently attracted much commentary and analysis~\cite{keener2009gospel}. The Sermon on the Mount is a spiritual discourse appearing in chapters 5-7 of the Book of Matthew of the New Testament. It was orally delivered by Jesus on an unidentified mountain commonly associated with the Mount of Beatitudes~\cite{macmillan1968atlas}. The Sermon on the Mount is an explanation from Jesus of how to live a life pleasing to God and what it means to be a Christian.  Jesus taught about subjects including prayer, salvation, justice, helping the poor, religious law, divorce, fasting, judging other people, etc. The Sermon on the Mount includes the Lord's Prayer and Beatitudes. It contains some of the most familiar Christian homilies and aphorisms.
 
 \subsection{Bible Translations}
  
 According to Wycliffe Bible Translators \cite{wycliffebibletranslators}, the King James Version of the Bible is available in 717 languages, the New Testament is available in an additional 1582 languages, and parts of the Bible are available in an additional 1196 languages. Although many translations use the original Greek text as the source language, others are translated from Latin ~\cite{wycliffe1388bible}, Aramaic 
 \cite{lamsa1933bible}, German~\cite{tyndale1526bible}, and mixtures of sources~\cite{matthew1537bible}. However,  within the same source language, different translations may be based on different variants of the text. For example, the King James Version \cite{carroll2008bible} and American Standard Version \cite{bible1995new} are both translated from Greek, but the former uses the Textus Receptus \cite{combs1996erasmus} whereas the latter relies on the Westcott \& Hort New Testament \cite{westcott1882new} and the Tregelles New Testament \cite{tregelles1857greek}.

\subsection{NLP for religious texts}

The use of NLP for religious texts is relatively new and gaining interest among researchers. 
Christodouloupoulos and Steedman \cite{christodouloupoulos2015massively} constructed a corpus of 100 Bible translations to serve as a resource for NLP applications, which has since been used extensively in machine translation research.
Carlson et al. \cite{carlson2018evaluating} evaluated style transfer algorithms on biblical texts. Garbhapu and Bodapati \cite{garbhapu2020comparative} implemented topic modelling and evaluated latent Dirichlet allocation and latent semantic analysis on a corpus of biblical texts. 
Xia et al~\cite{xia-yarowsky-2017-deriving} jointly analysed 27 versions of the English Bible, creating a consensus alignment of speech tokens between all versions rather than between individual pairs. This led to a richer set of features, such as part-of-speech tags and headwords rather than typical pairwise analysis. 
Ebrahimi and Kann~\cite{ebrahimi2021adapt} evaluated zero-shot learning of pre-trained language models to low-resource languages where Bible translations were the only data available to use as a basis for adaptation. Leveraging the Bible as a resource led to a significant improvement in tasks such as part-of-speech tagging and named entity recognition.

In contrast, the studies that go the other way and use NLP as a tool for biblical scholarship are less common. Nakov et al~\cite{nakov2000latent} implemented latent semantic analysis on a dataset of  monotheisticsacred texts. The statistical method was able to recover the high level of similarity between texts from the same religion, between the Old and New Testaments, and between Biblical and Quranic texts.
Franklin \cite{franklin2018differences} evaluated the emotionality of different English translations using an association lexicon method.
Putniņš et al \cite{putnicnvs2006advanced} evaluated a number of authorship attribution algorithms on a corpus of biblical texts and applied them to specific chapters of disputed authorship, which added powerful new evidence to the ongoing debate.
Popa et al. \cite{popa2019extracting} applied ontology learning to identify the most important concepts in the Bible and the relationships between them.
Hu \cite{hu2012unsupervised} performed topic modelling on the \textit{Book of Psalms} and \textit{Book of Proverbs}, both of which are collections of teachings, prayers, and poems on various subjects.
Yeshurun and Kimelfeld \cite{yeshurun2021extracting} developed a pipeline for identifying biblical quotations in historical religious texts.
Bader et al. \cite{bader2011multilingual} performed translation-free sentiment analysis using Bible data in multiple languages where only one language had semantic tagging available.
Jones et al.~\cite{jones2022machine} evaluated several supervised learning algorithms on the task of reconstructing the original text of the Greek New Testament. They found high agreement between the learning algorithms and human experts as well as between the different algorithms.
Varghese and Punithavalli~\cite{varghese2019lexical} used text-mining techniques to perform lexical and semantic analysis of religious books used by Jews, Christians, and Muslims. They compute polarity, subjectivity, and various similarity scores.
Mohamed and Shokry~\cite{mohamed2022QSST} used word embedding to perform information retrieval on the Quran. By using cosine similarity between the learned features of both topics and queries, the model was able to closely match human annotators, retrieving relevant verses with high precision.

\section{Methodology}

\subsection{Data: Bible - Sermon of Mount}

Although there are hundreds of  English translations of the Bible, we focus on selected prominent versions that are mostly used. According to a study ~\cite{bibleinamericanlife} by researchers at Indiana University and Purdue University, among respondents who read the Bible in the United States, 55\% read the King James Version, 19\% read the New International Version, 7\% read the New Revised Standard Version, 6\% read the New American Bible, 5\% read the Living Bible, and 8\% read other translations. Therefore, our analysis includes the three most popular English translations in the United States. 

Hence, we select the King James Version (KJV), the New International Version (NIV), and the New Revised Standard Version (NRSV) as our key translations. Apart from these, we select the Lamsa Version (LV)  because it has been translated from Aramaic rather than from Greek. Finally, we include the  Basic English Version (BEV), which uses a restricted lexicon of simple English words. In each of these translations, the Sermon on the Mount consists of three chapters divided into the same number of verses, but the verses themselves differ significantly. Table  1  presents a summary of the number of words and verses in each version of the Sermon of the Mount from the Bible.

\begin{table*}[htbp!]
\begin{center}
\begin{tabular}{|l|l|l|l|l|l|l|} 
\hline
 \textbf{Chapter} &  \textbf{Verses} & \textbf{Words-KJV}& \textbf{Words-NIV}& \textbf{Words-NRSV} & \textbf{Words-LV} & \textbf{Words-BEV} \\ 
 \hline
 5 &48&5221&5392&5105&5586&5584\\ 
 \hline
 6&34&3807&4008&3807&4148&4216\\ 
 \hline
 7&29&2945&3052&2848&3120&3194\\ 
 \hline
\end{tabular}
\caption{Verse and word count for respective translations}
\label{tab:word_count}
\end{center}

\end{table*}

 \subsection{Large language models for  sentiment analysis }
 

The long-short-term memory (LSTM) network is a class of recurrent neural networks (RNNs) that  has prominently been used for processing temporal  data and developing language models due to their capability to model temporal sequences  \cite{hochreiter1997long}. The LSTM network was developed for addressing model training that emerges due to learning of long-term dependency problems in conventional RNNs \cite{hochreiter1998vanishing}. Since then, LSTM networks have been used in a wide range of applications that include processing of text, speech, time series and image data \cite{yu2019review}. Several major advancements in LSTM networks have taken place that include developing hybrid variants with deep learning models such as conventional neural networks and bi-directional and encoder-decoder architectures \cite{van2020review}. To address certain limitations motivated by cognitive science, Wang et al. \cite{wang2016attention}  developed an attention-based LSTM network. The attention mechanism gives more attention to important information, such as specific words in a sentence, that enables it to provide better language modelling capabilities as demonstrated by the encoder-decoder LSTM  with attention, which is also known as a \textit{transformer}  \cite{vaswani2017attention}. The transformer model for language tasks features millions of trainable parameters, which is a computationally challenging task. A robust model requires a large corpus of data. This led to the development of a pre-trained model known as BERT\cite{devlin2018bert}. Hence, developers and researchers can utilize and refine the pre-trained  BERT model for language modelling tasks rather than developing a model from nothing. 
BERT learns contextual relations between words and sub-words in a corpus.

 BERT enables the understanding of ambiguous text by using surrounding text to establish context. BERT has been applied in a wide range of NLP tasks that include semantic and sentiment analysis, text prediction, text generation, text summarization, text responses, and polysemy resolution \cite{koroteev2021bert}. A wide range of implementations of BERT exist that address computational challenges  and  domain-specific applications \cite{zhou2023comprehensive} such as law and medicine \cite{
chalkidis2020legal,rasmy2021med}. BERT has been widely used for sentiment analysis tasks for different domains \cite{acheampong2021transformer}. Hence, it is suitable for our application. The sentence-BERT (S-BERT)\cite{reimers2019sentence} model improves the BERT model by reducing the computational time to derive semantically meaningful sentence embedding. Hence, we will use S-BERT in our framework for sentiment analysis.


\subsection{Framework}

We present the framework that compares different English translations (versions) of the Bible focusing on the Sermon on the Mount using sentiment analysis. We examine several versions to evaluate their similarity and differences given differences in their sources, the translation style, and the time period of translation. 
Sentiment analysis refers to the processing of human emotions in text using an NLP model. For example, text may be “optimistic,” “pessimistic,” “anxious,” “thankful,” etc. Sentiment analysis can feature single-label (only one sentiment) or multi-label (multiple sentiments) analysis. Our study focuses on multi-label sentiment analysis where a sentence can be both "pessimistic" and "anxious." Our framework features S-BERT model, which is refined using the Senwave dataset for multi-label sentiment analysis. We also use the same model for sentiment polarity analysis where we present sentiment scores rather than sentiment type. 

\begin{figure*}[htbp]
    \centering
    \includegraphics[scale=0.55]{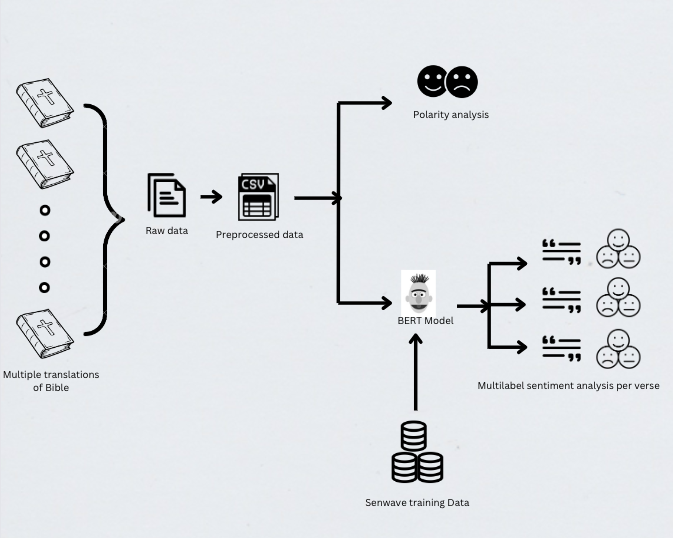}
    \caption{Framework diagram showing major components that include preprocessing and sentiment analysis across translations of Sermon on the Mount. }
    \label{fig:framework_diagram}
\end{figure*}

In our framework shown in Figure  1, we first convert the text files to CSV file format which features contain each chapter with its verses. We then removed stop words (using Natural Language Toolkit \footnote{\url{https://www.nltk.org/}}) and grouped together inflected forms of a word to be analyzed as a single item (lemmatisation). We also normalized to UTF-8\cite{yergeau2003utf} and processed split words at punctuation using regular expressions \cite{karttunen1996regular}. We implement this for the selection versions of the Bible's Sermon on the Mount summarised in Table 1.

Note that we have two types of sentiment analysis that are evaluated by our framework: sentiment polarity score and multi-label sentiment classification. 
We use the  AFINN sentiment analysis \cite{nielsen2011new} for sentiment polarity score.  AFINN is a database that features a  set of English words rated for sentiment polarity with an integer
between  -5 and +5 manually labelled by Finn {\AA}rup Nielsen in 2009-2011 \cite{nielsen2011afinn}. In the case of sentiment classification, we fine-tune the S-BERT model using Senwave data \cite{yang2020senwave}, which features 10,000 tweets hand labelled by 50 experts with 10 different sentiments plus 1 label that refers to the COVID-19 official report. We limited to 10 sentiments and removed the “official report” in data processing and utilized the setup of our framework from previous work   \cite{chandra2022semantic} that was used for translation analysis of the Bhagavad Gita, the central text of Hinduism.

In our framework, we pass each version of the Sermon on the Mount, chapter by chapter, and record the sentiment classification for every verse along with the polarity score. We then provide an analysis by comparing the sentiment classification levels and polarity scores across the different chapters for respective versions of the Sermon on the Mount.

 An \emph{n-gram} \cite{kumar2017introduction} is typically used to provide basic statistics on a document using a continuous sequence of words and elements.  A  \emph{bi-gram} is a sequence of two words and a \emph{tri-gram} is a sequence of three words. We also provide a bi-gram and tri-gram analysis to evaluate the difference in the vocabulary used by the respective versions.

\section{Results}

\subsection{Data analysis}

\begin{figure}
     \centering
     \begin{subfigure}[b]{0.5 \textwidth}
         \centering
         \includegraphics[width=\textwidth]{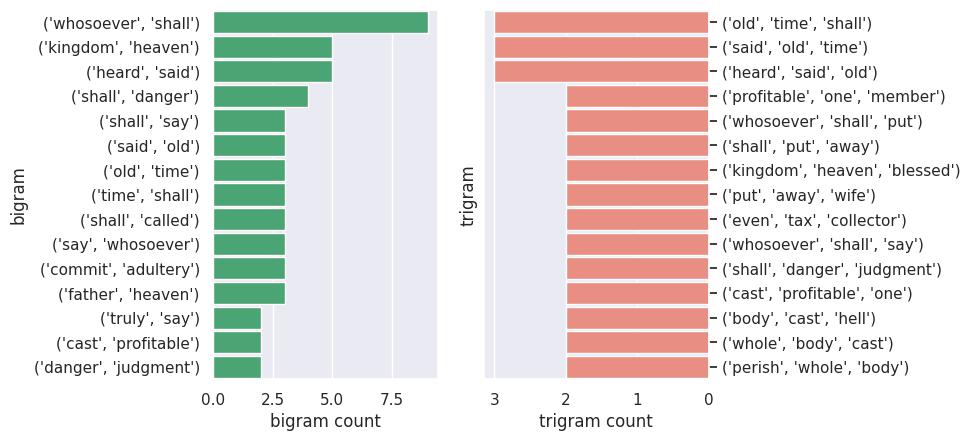}
         \caption{KJV} 
     \end{subfigure}
     \hfill
     \begin{subfigure}[b]{0.5 \textwidth}
         \centering
         \includegraphics[width=\textwidth]{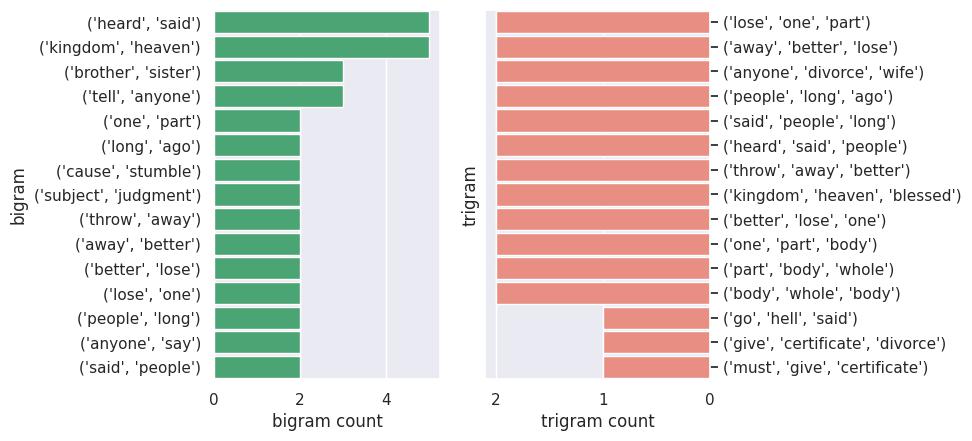}
         \caption{NIV} 
     \end{subfigure} 
       \hfill
     \begin{subfigure}[b]{0.5 \textwidth}
         \centering
         \includegraphics[width=\textwidth]{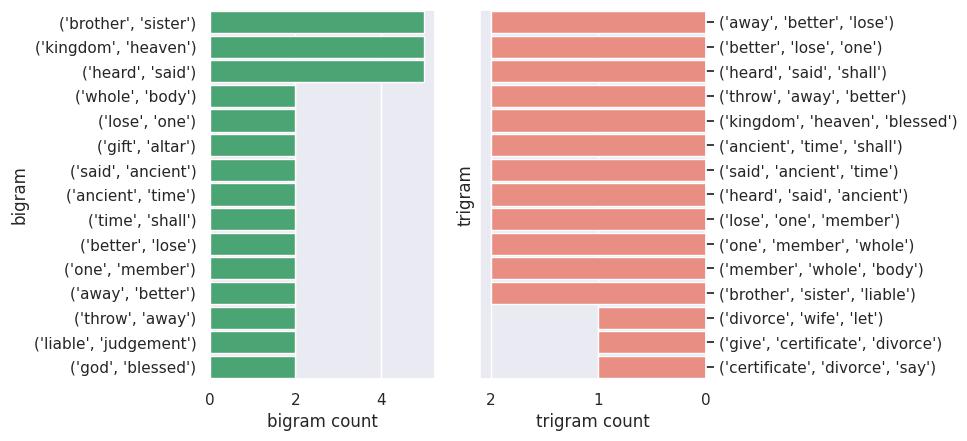}
         \caption{NRSV} 
     \end{subfigure} 
 \caption{Visualisation of top 10 bi-grams and tri-grams for the selected versions of Sermon on the Mount.}
      \label{fig:bitri}
\end{figure}

\begin{figure}
     \centering
     \begin{subfigure}[b]{0.5 \textwidth}
         \centering
         \includegraphics[width=\textwidth]{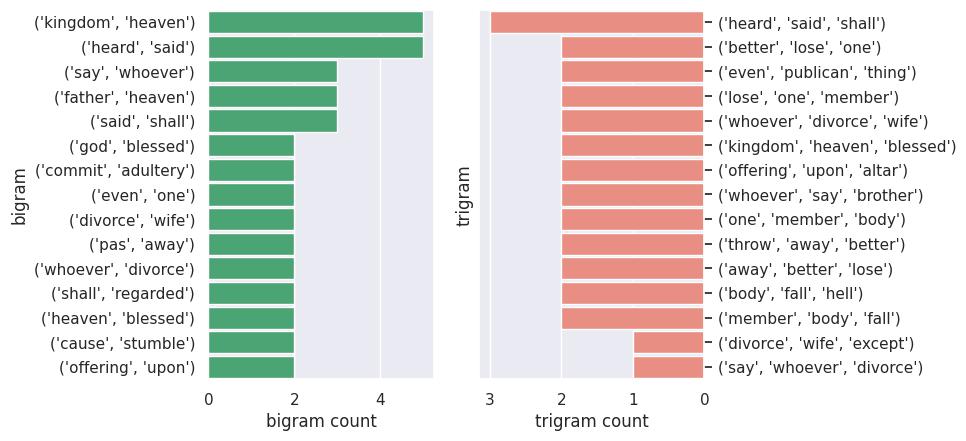}
         \caption{LV} 
     \end{subfigure}
     \hfill
     \begin{subfigure}[b]{0.5 \textwidth}
         \centering
         \includegraphics[width=\textwidth]{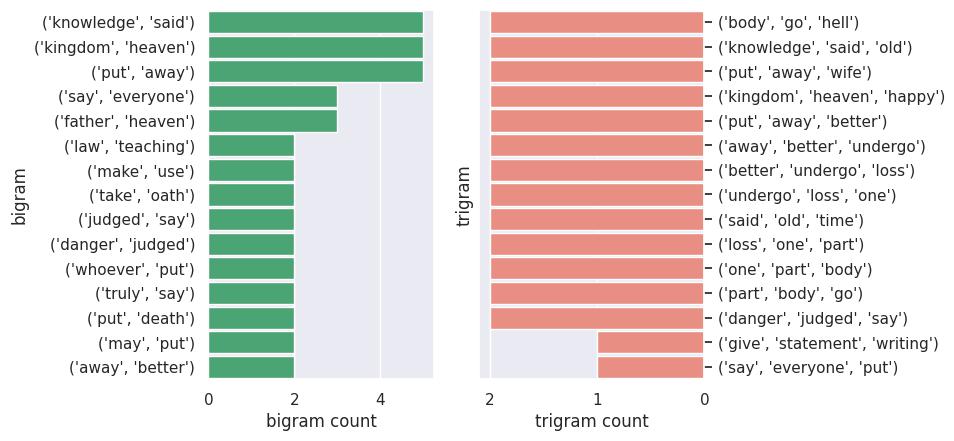}
         \caption{BEV} 
     \end{subfigure}  
 \caption{Visualisation of top 10 bi-grams and tri-grams for the selected versions of Sermon on the Mount.}
      \label{fig:bitri2}
\end{figure}

We begin by analyzing the top 10 bi-grams and tri-grams for each translation as shown in Figure \ref{fig:bitri} (KJV, NIV, and NRSV), Figure \ref{fig:bitri2} (LV  and BEV). We observe that the repeated tri-grams are rare across all versions with most having no tri-grams that appear more than twice and none having a tri-gram that appears more than $3$ times. This indicates a low degree of repetition and a significant difference in the vocabulary across the versions. Among bi-grams, the only one that consistently appears with high frequency is  ['kingdom', 'heaven'], which is likely reflecting the construction "Kingdom of Heaven", suggesting that this concept persists regardless of style, source language, and time period.

\subsection{Sentiment Analysis}

 We use the BERT model from our framework (Figure 1) for verse-by-verse sentiment analysis of the respective Sermon on The Mount translations.

 \begin{figure}[htbp!]

    \includegraphics[scale=0.2]{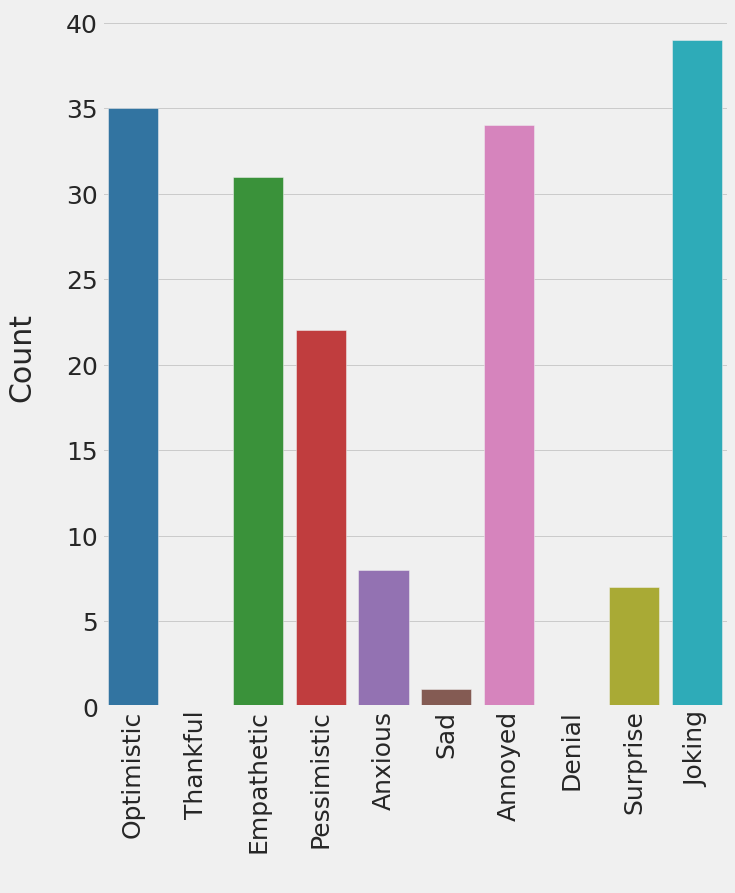}
    \caption{Cumulative sentiments of the respective chapters. }
    \label{fig:cumul_sa}
    
\end{figure}

We present the cumulative sentiments from the selected chapters of the KJV as it's the most popular translation.  In Figure \ref{fig:cumul_sa}, surprisingly \textit{joking} is the sentiment detected the most from the verses, which is not true as Sermon on the Mount is a religious text with a serious tone. However, as some of the teachings are metaphorical, our model detects them as joking. We note that the joking label does not imply a joke but suggests that the verses feature some level of humor. We revisit the famous parable of the \text{Mote and the Beam} (Chapter 7: verses 1-5) where Jesus warns his followers of the dangers of judging others and states that they too would be judged by the same standards. In Table \ref{tab:four}, in Chapter 7 verse 5, we find that KVJ, NIV and NRSV are classified as annoyed and joking while LV and BEV classified the verse as joking. As well as the injunction to "turn the other cheek" (Chapter 5: verse 39) is also interpreted as joking in all the translations.

In Figure 7, we observe that "optimistic" is the second most expressed sentiment in the teachings followed by "annoyed", "empathetic" and "pessimistic" in that order. These results comport with what we might expect: the text is largely optimistic and teaches that entrance to heaven is within reach for all people (hence optimism and empathy), but also includes verses that condemn injustice and inequity (reflecting annoyance and pessimism).
Notable in their absence are sentiments such as thankfulness, sadness, and denial. Indeed, since the sermon is a generous offer or promise being extended to the congregation, there is no cause for Jesus to be thankful or sad or to deny any accusation.

\begin{table*}[htbp!]
\footnotesize
\begin{center}
\begin{tabular}{p{0.1\textwidth}p{0.15\textwidth}p{0.15\textwidth}p{0.15\textwidth}p{0.15\textwidth}p{0.15\textwidth}} 
\hline
 \textbf{} &\textbf{King James} &  \textbf{New International} & \textbf{New Revised Standard}& \textbf{Lamsa}& \textbf{Basic English} \\ 
 \hline
Verse-5 Chapter-7&you hypocrite first take out the beam out of your own eye and then shall you see clearly to take out the speck out of your brother s eye.
&you hypocrite first take the plank out of your own eye and then you will see clearly to remove the speck from your brother s eye.
&you hypocrite first take the log out of your own eye and then you will see clearly to take the speck out of your neighbour s eye.
&o hypocrites first take out the beam from your own eye and then you will see clearly to get out the splinter from your brother s eye.
&you false one first take out the bit of wood from your eye then will you see clearly to take out the grain of dust from your brother s eye.

\\  \hline
Sentiments&Annoyed, Joking&Annoyed, Joking&Annoyed, Joking&Joking&Joking
\\  \hline

 Verse-34 Chapter-6&take therefore no worry for the tomorrow for the tomorrow shall take worry for the things of itself. sufficient to the day is the evil thereof.
&therefore do not worry about tomorrow for tomorrow will worry about itself. each day has enough trouble of its own.
&so do not worry about tomorrow for tomorrow will bring worries of its own. today s trouble is enough for today.
&therefore do not worry for tomorrow for tomorrow will look after its own. sufficient for each day is its own trouble.
&then have no care for tomorrow tomorrow will take care of itself. take the trouble of the day as it comes.

\\  \hline
Sentiments&Pessimistic&Pessimistic, Sad, Annoyed&Optimistic, Anxious&Optimistic, Empathetic&Optimistic\\  \hline

 Verse-39 Chapter-5&but i say to you that you resist not evil but whosoever shall strike you on your right cheek turn to him the other also.
&but i tell you do not resist an evil person. if anyone slaps you on the right cheek turn to them the other cheek also.
&but i say to you do not resist an evildoer. but if anyone strikes you on the right cheek turn the other also
&but i say to you that you should not resist evil but whoever strikes you on your right cheek turn to him the other also.
&but i say to you do not make use of force against an evil man but to him who gives you a blow on the right side of your face let the left be turned.

\\  \hline
Sentiments&Joking&Annoyed, Joking&Annoyed, Joking&Joking&Annoyed, Joking\\  \hline
Verse-6 Chapter-7&give not that which is holy to the dogs neither cast you your pearls before pigs lest they trample them under their feet and turn again and tear you.&do not give dogs what is sacred do not throw your pearls to pigs. if you do they may trample them under their feet and turn and tear you to pieces.&do not give what is holy to dogs and do not throw your pearls before swine or they will trample them under foot and turn and maul you.&do not give holy things to the dogs and do not throw your pearls before the swine for they might tread them with their feet and then turn and rend you.&do not give that which is holy to the dogs or put your jewels before pigs for fear that they will be crushed under foot by the pigs whose attack will then be made against you.\\\hline
Sentiments&Annoyed, Joking&Annoyed, Joking&Annoyed, Joking&Joking&Anxious, Annoyed, Joking\\\hline
Verse-31 Chapter-6 &therefore take no worry saying what shall we eat or what shall we drink or wherewithal shall we be clothed&so do not worry saying what shall we eat or what shall we drink or what shall we wear&therefore do not worry saying what will we eat or what will we drink or what will we wear&therefore do not worry or say what will we eat or what will we drink or with what will we be clothed&then do not be full of care saying what are we to have for food or drink or with what may we be clothed\\\hline
Sentiments&Optimistic, Anxious&Surprise&Surprise&Empathetic&Pessimistic, Annoyed\\\hline
Verse-31 Chapter-5&it has been said whosoever shall put away his wife let him give her a writing of divorcement&it has been said anyone who divorces his wife must give her a certificate of divorce.&it was also said whoever divorces his wife let him give her a certificate of divorce.&it has been said that whoever divorces his wife must give her the divorce papers.&again it was said whoever puts away his wife has to give her a statement in writing for this purpose\\\hline
Sentiments&Joking&Joking&Joking&Pessimistic, Anxious, Annoyed, Joking&Annoyed, Joking\\\hline
Verse-5 Chapter-6&and when you pray you shall not be as the hypocrites are for they love to pray standing in the synagogues and in the corners of the streets that they may be seen of men. truly i say to you they have their reward.&and when you pray do not be like the hypocrites for they love to pray standing in the synagogues and on the street corners to be seen by others. truly i tell you they have received their reward in full.&and whenever you pray do not be like the hypocrites for they love to stand and pray in the synagogues and at the street corners so that they may be seen by others. truly i tell you they have received their reward.&and when you pray do not be like the hypocrites who like to pray standing in the synagogues and at the street corners so that they may be seen by men. truly i say to you that they have already received their reward.&and when you make your prayers be not like the false hearted men who take pleasure in getting up and saying their prayers in the synagogues and at the street turnings so that they may be seen by men. truly i say to you they have their reward.\\\hline
Sentiments&Optimistic, Empathetic&Optimistic, Empathetic&Optimistic, Empathetic&Optimistic, Joking&Empathetic, Annoyed\\\hline
\end{tabular}
\caption{Sentiments across versions}
\label{tab:four}
\end{center}

\end{table*}

\begin{figure}[htbp]
\centering
    \includegraphics[scale=0.3]{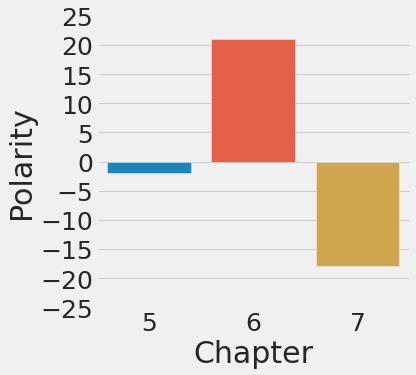}
    \caption{Polarity Of Jesus's Speech}
    \label{fig:polarity}
    
\end{figure}  

\begin{figure}[htbp]
    \includegraphics[scale=0.35]{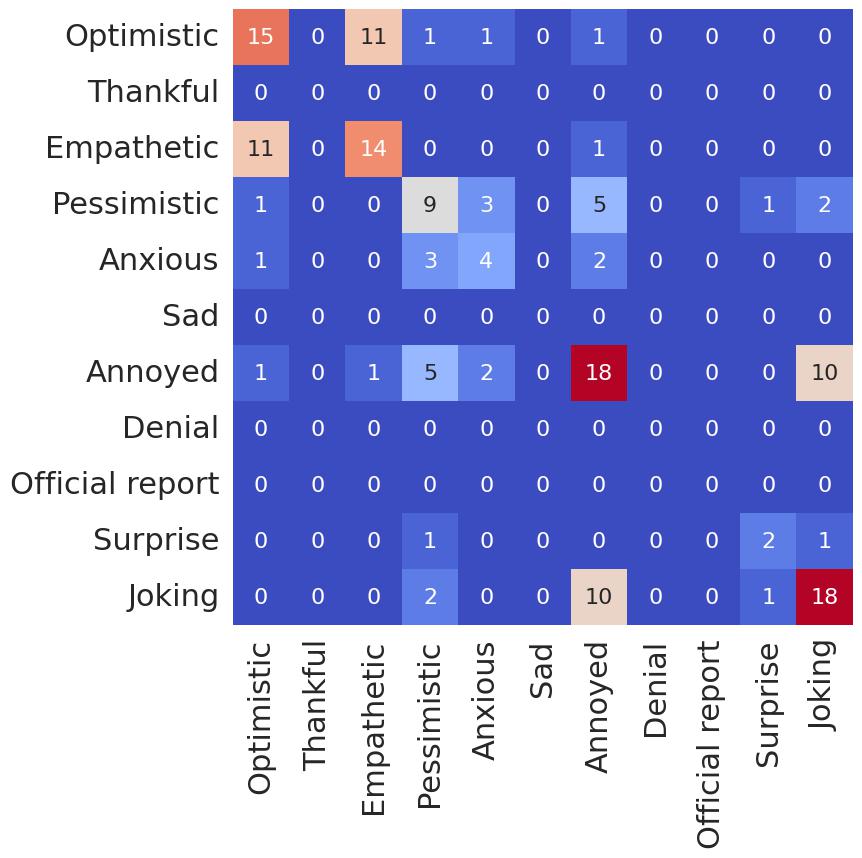}
    \caption{Heatmap, King James Version-Chapter5 Verse1}
    \label{fig:chap5_hmap}
\end{figure}

\begin{figure}
     \centering
     \begin{subfigure}[b]{0.4 \textwidth}
         \centering
         \includegraphics[width=\textwidth]{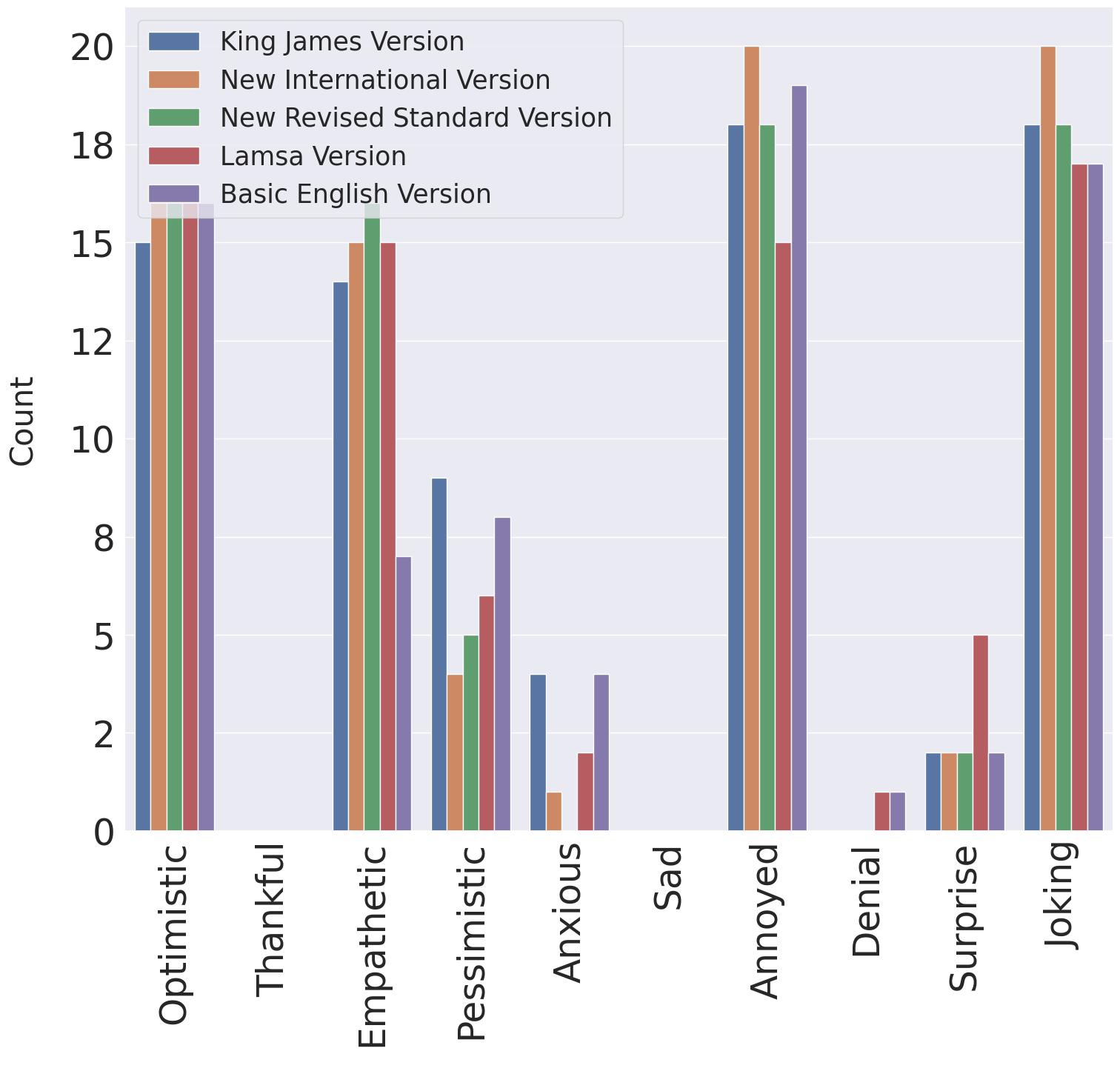}
         \caption{Chapter 5} 
     \end{subfigure}
     \hfill
     \begin{subfigure}[b]{0.4 \textwidth}
         \centering
         \includegraphics[width=\textwidth]{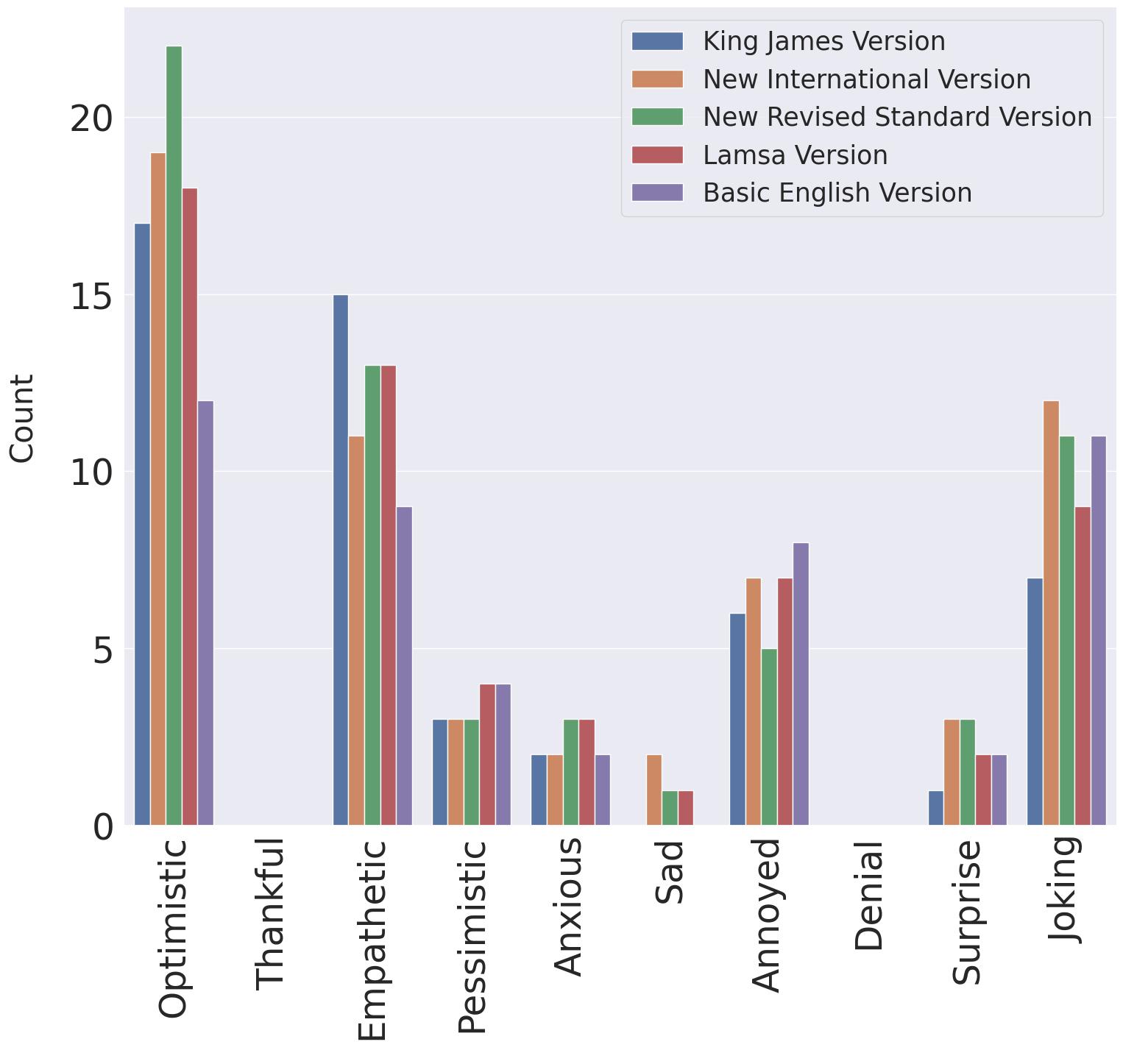}
         \caption{Chapter 6} 
     \end{subfigure} 
       \hfill
     \begin{subfigure}[b]{0.4\textwidth}
         \centering
         \includegraphics[width=\textwidth]{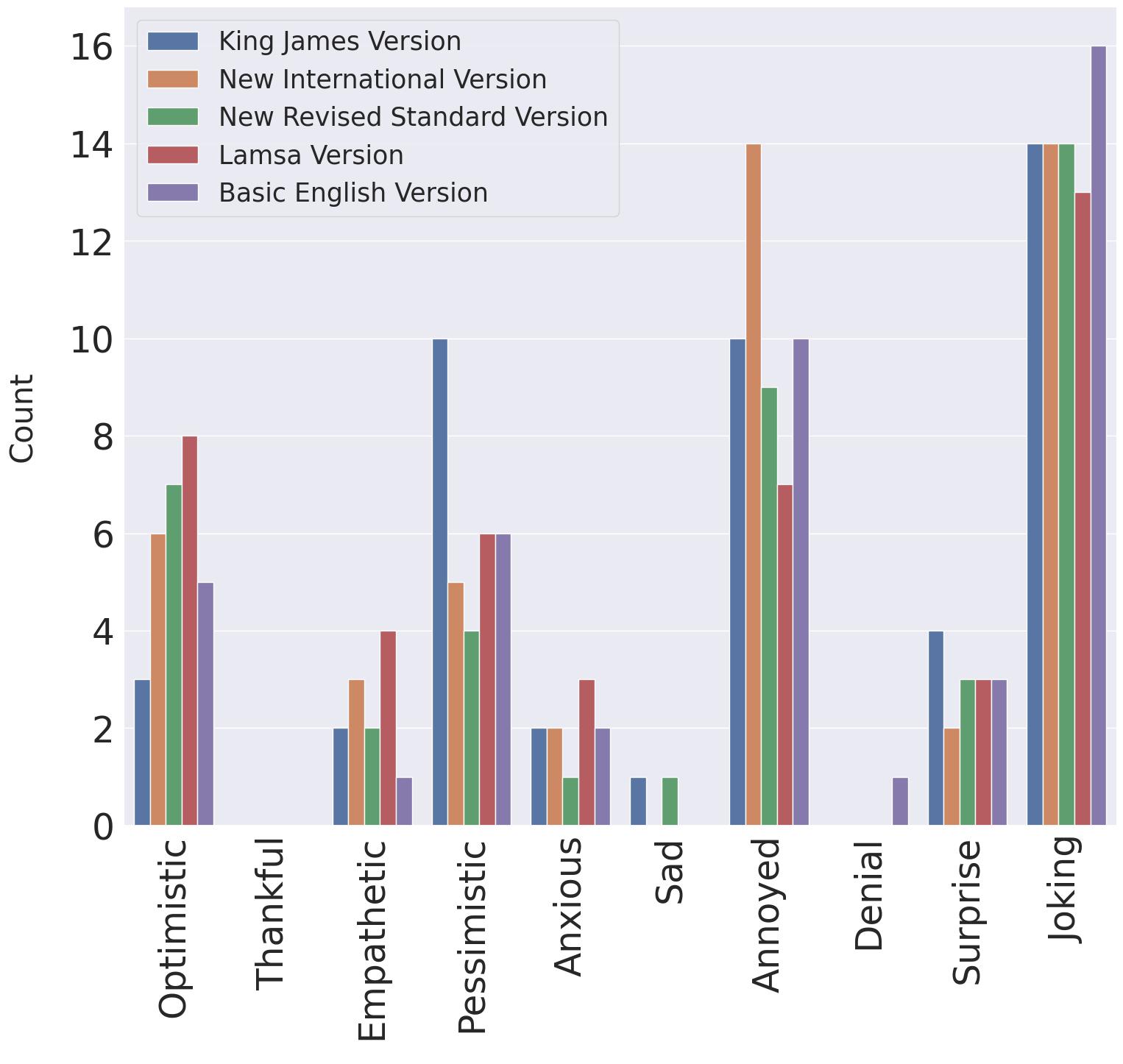}
         \caption{Chapter 7} 
     \end{subfigure} 
 \caption{Chapter-wise Sentiment Analysis for KJV of Sermon On the Mount.}
      \label{fig:senti}
\end{figure}

\begin{figure}
     \centering
     \begin{subfigure}[b]{0.5\textwidth}
         \centering
         \includegraphics[width=\textwidth]{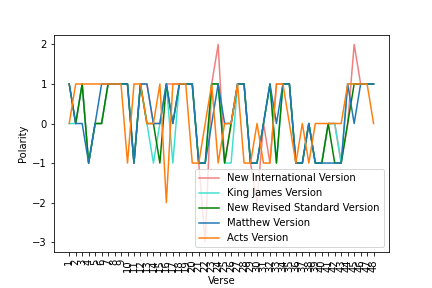}
         \caption{Chapter 5} 
     \end{subfigure}
     \hfill
     \begin{subfigure}[b]{0.5 \textwidth}
         \centering
         \includegraphics[width=\textwidth]{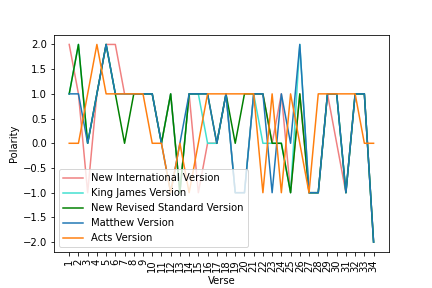}
         \caption{Chapter 6} 
     \end{subfigure} 
       \hfill
     \begin{subfigure}[b]{0.5 \textwidth}
         \centering
         \includegraphics[width=\textwidth]{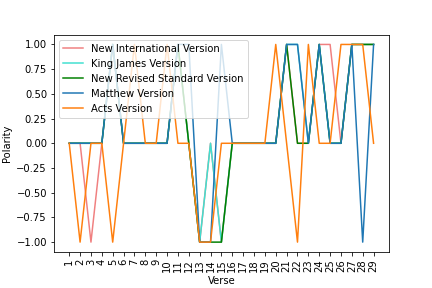}
         \caption{Chapter 7} 
     \end{subfigure} 
 \caption{Sentiment polarity chapter-wise for KJV of Sermon On the Mount.}
      \label{fig:polarity}
\end{figure}

Figure \ref{fig:polarity} presents the chapter-wise sentiment analysis for various translations. We find that "optimistic," "joking," and "annoyed" are the top three sentiments expressed in Chapter 5 while Chapter 6 has more optimistic and empathetic sentiments than annoyed and joking. Chapter 7 has the most joking sentiments followed by annoyed and optimistic. It is interesting to note that KJV has significantly more pessimistic and less optimistic sentiments for Chapter 7 when compared to other versions. In Chapter 5, we find that KJV and NRSV have similar sentiment counts that differ significantly in Chapters 6 and 7.  We notice that in Chapter 5, none of the versions had a "sad" label and in Chapter 6, BEV had no label detected for the same sentiment. In Chapter 7, only KJV and NRSV detected the "sad" sentiment label. We find that the "denial" label is only present in LV and BEV in Chapter 5 and Chapter 6 has not detected this sentiment while in Chapter 7, only BEV detected the sentiment.  We note that "annoyed" sentiment covers that Jesus seems to be annoyed at the things evil and foolish people do. We also note that the "thankful" label has not been detected across all of the chapters and versions. This may imply that Jesus was not openly being thankful to people for listening to them as typically done in political speeches. His goal was to impart spiritual and philosophical ideals and not gain political support.

We also provide an analysis of how Jesus conveyed his ideals in speech based on sentiment polarity (AFINN score) throughout the 3 chapters (Figure \ref{fig:polarity}). A higher and positive score implies more positive sentiments such as optimism, thankful, and empathy. The lower and negative score implies negative sentiments such as denial, sadness, and pessimism.

We can see in Figure \ref{fig:polarity} that Jesus speaks with an overall negative sentiment score in the 5 and 7 Chapters (-2 and -18), and with a high positive sentiment score (21) in Chapter 6. We notice a high fluctuation in Chapter 6 verses 21-28 in KJV. This segment compares material goods with spiritual goods so jumps in polarity make sense.
As we can see, a similar trend is followed in all of the three versions. In some cases, the KJV and the NRSV have very similar scores and the scores for the NIV are slightly different, but follow the same trend.  


\section{Discussion}

Our study focused on translations of the Bible to English and sentiment analysis on chapters comprising the Sermon on the Mount.  Our results in general indicate that large language models for sentiment analysis can help us gain insight into how the different translations affect the sentiments portrayed in the resulting text.  Our results demonstrate the capabilities of our framework to explore the similarities and differences between different translations.  In summary, on the one hand, we see a general agreement among the translations as the variance in any particular sentiment is not large (Figure 7). On a finer scale, however, we see that some translations are outliers in certain sentiments, such as the KJV evincing significantly less optimism and more pessimism than the NRSV. In terms of sentiment polarity (Figure 5), Chapter 5 features an overall neutral polarity, Chapter 6 has the most positive sentiment polarity, and Chapter 7 has the highest negative polarity. 

 The parable of the Mote and the Beam as well as the exhortation to turn the other cheek were both identified as jokes or having some level of humor in all translations. Furthermore, we note that "thankful" was not detected in any chapter across those translations, which may imply that Jesus was not giving a political speech to gain followers, but a sermon or a spiritual/philosophical discussion that used some humour while mostly being optimistic (Chapter 5 and 6) and annoyed and humorous (Chapter 7). This indicated that while being annoyed, he tried to ease out the discussion with a humorous tone to create a less stressful situation arising from the discussions 

 It is well known that translations can vary in time, location, and religious denomination \cite{strauss2010distorting}. Hence, the translations reflect the prejudices of the translator, their agenda, the political circumstances of the time, and the broader surrounding culture \cite{mojola2014old}. For example, the Luther Bible \cite{lobenstein2017martin} was intentionally translated into contemporary vernacular rather than the literary prose of the original. This reflects Martin Luther's desire to weaken the control of the Catholic church by allowing everyday people to read and understand the Bible without intermediaries~\cite{dickens1974luther}.  As with the rest of the New Testament, the Sermon on the Mount was originally written in Koine Greek. However, some scholars argue that various parts were originally Aramaic \cite{stuckenbruck1991approach} and have since been translated many times. William Chamberlin categorized approximately 900 English translations of the Bible in ~\cite{chamberlin1991catalogue}.

In terms of limitations, our study also highlights an important shortcoming with modern sentiment analysis techniques:  the tendency to interpret metaphors incorrectly as jokes (humour). The Bible is often heavily laden with metaphors, and this is reflected by the high number of verses that were identified as joking. We also highlight that a further limitation is that we reviewed only English translations without reviewing the source language. Translations, especially metaphors, are often mistranslated and are prone to be affected by the biases of the translator, which depend on the  time period and cultural settings. Finally, we note that our sentiment classification model was refined by the Senwave dataset, which was manually labelled for COVID-19 tweets worldwide. In our earlier study \cite{chandra2021covid}, we reviewed the bi-grams and tri-grams of the tweets, which indicate that the vocabulary of the training dataset is significantly different from the respective versions of the Sermon of the Mount in this study (Figures 2 and 3). The difference in the vocabulary indicates a bias in sentiment labelling. We also note that the timeline and the nature of COVID-19 and associated discussions are very different when compared to the Sermon on the Mount. In future works, we would need to manually label related texts to remove the sentiment labelling bias in the training data. 

In future work, our framework can be straightforwardly extended to other translations from which we could then identify outliers in the same way we detected that the KJV is an outlier for pessimism (Figure 7, Panel c). Future work can focus on incorporating explicit metaphor detection \cite{tsvetkov2014metaphor,song2021verb} in the analysis\ to capture the full depth of such verses. The study can also be extended to related Biblical texts along with texts about interpretations of the Sermon on the Mount \cite{kissinger1975sermon}.  Topic modelling has recently been done in the case of Hindu texts such as the Bhagavad Gita and the Upanishads \cite{chandra2022artificial}, which further motivates extension to the Bible and related texts. Hence, topic modelling on different translations of the Sermon on the Mount and other parts of the Bible can be considered. Through topic modelling, the Sermon on the Mount can be compared with other earlier texts such as the Dhammapada of Buddha \cite{easwaran2007dhammapada}, the Upanishads, and the Bhagavad Gita.

 \section{Conclusions}

 We presented a sentiment analysis framework for comparing selected translations of Sermon on the Mount. Our results highlighted the varying sentiments across the chapters and verses associated with different themes that link with varying tones and styles used by Jesus to convey his teachings. We also find that the vocabulary of the respective translations was significantly different, which may be associated with the timeline and cultural setting of the translation. We found that a certain level of humor was detected in all the chapters along with different levels of optimism and empathy, which form the core teachings of Sermon on the Mount. Our framework can be extended to other languages and related texts, which can be helpful in evaluating the quality of translations.


\section*{Code and Data}
Code and Data \url{https://github.com/sydney-machine-learning/sentimentanalysis-Bible}

\bibliographystyle{elsarticle-num} 
\bibliography{ref.bib} 
\end{document}